\documentclass{article}
\usepackage{spconf,amsmath,graphicx}

\usepackage{amssymb,amsfonts}
\usepackage{multirow}
\usepackage[capitalise]{cleveref}
\usepackage{caption}
\usepackage{subcaption} 
\usepackage{siunitx}
\usepackage{xurl}

\usepackage{flushend}

\title{
A Novel Local-Global Feature Fusion Framework for Body-weight Exercise Recognition with Pressure Mapping Sensors
}
\def\correspondingauthor{\sthanks{Corresponding author: sungho.suh@dfki.de}}

\name{Davinder Pal Singh $^{1}$\sthanks{D. Singh and L. Ray - These authors contributed equally to this work.}, Lala Shakti Swarup Ray $^{2*}$, \ Bo Zhou$^{1,2}$, \ Sungho Suh$^{1,2}$\correspondingauthor, \ Paul Lukowicz$^{1,2}$
}
\address{$^{1}$ Department of Computer Science, RPTU Kaiserslautern-Landau, Kaiserslautern, Germany\\
	$^{2}$ German Research Center for Artificial Intelligence (DFKI), Kaiserslautern, Germany}

\begin{document}
%
\maketitle
\begin{abstract}
We present a novel local-global feature fusion framework for body-weight exercise recognition with floor-based dynamic pressure maps. One step further from the existing studies using deep neural networks mainly focusing on global feature extraction, the proposed framework aims to combine local and global features using image processing techniques and the YOLO object detection to localize pressure profiles from different body parts and consider physical constraints. The proposed local feature extraction method generates two sets of high-level local features consisting of cropped pressure mapping and numerical features such as angular orientation, location on the mat, and pressure area. In addition, we adopt a knowledge distillation for regularization to preserve the knowledge of the global feature extraction and improve the performance of the exercise recognition. Our experimental results demonstrate a notable 11 percent improvement in F1 score for exercise recognition while preserving label-specific features.

\end{abstract}
\section{Introduction}
\label{sec:intro}

    Human activity recognition (HAR) identifies human movements based on observations of the person and the environment by using input signals from videos or various on-body/environmental sensors. HAR has been applied in surveillance systems, behavior analysis, fitness tracking, healthcare systems, and industrial optimization applications \cite{prati2019sensors, liaqat2019wearbreathing}. Among various HAR applications, sports activity recognition is also a well-studied topic in mobile, wearable, and ubiquitous computing fields to assist people in tracking their activity levels and achieving their fitness goals \cite{mencarini2019designing, cust2019machine, rajvsp2020systematic}. Various fitness-tracking applications have been developed using inertial measurement unit (IMU) sensors in mobile devices such as smartphones and smartwatches. In addition, some fitness equipment such as treadmills, weight training equipment, and cycling bikes provide user's real-time workout tracking information and guidance based on the integrated sensors. Recently, due to the global pandemic of COVID-19, people's lifestyles have changed, and exercise at home has become a trend. According to a recent survey \cite{wilke2020restrictercise}, body-weight exercises are among the most popular home workouts because of the flexible choices of exercise types, location, and low cost. Wearable device-based and computer vision-based methods have been proposed to recognize body-weight exercises. The wearable device-based method \cite{kwon2021approaching} showed that some types of body-weight exercises could be recognized, but they require more on-body sensors for more accurate and comprehensive monitoring. Computer vision-based methods can extract and recognize poses and motions from typical upright positions, but they can be degraded by visual occlusion, and poor light conditions \cite{garbett2021towards}. In addition, they are subject to privacy concerns for applications inside private homes. 
    
    As an alternative to these limitations, Sundholm et al. \cite{sundholm2014smart} proposed a gym exercise recognition system with unobtrusive pressure mapping sensors. The system detects ten full-body exercises. The pressure mapping sensors have been applied in various applications such as monitoring driver motions in a car seat, nutrition monitoring \cite{cheng2016smart}, in-bed sleep posture recognition \cite{diao2021deep}, and human and robot position recognition for human-robot collaboration in an intelligent factory manufacturing line \cite{peter2020object}. Recently, in our previous work \cite{zhou2022quali}, we proposed a method to detect the quality of execution during body-weight exercises and demonstrated an effective 3D convolutional neural networks (3DCNN) model outperforms more complex state-of-the-art machine learning models. 
    
    \begin{figure*}[!t]
        \begin{center}
        \includegraphics[width=\linewidth]{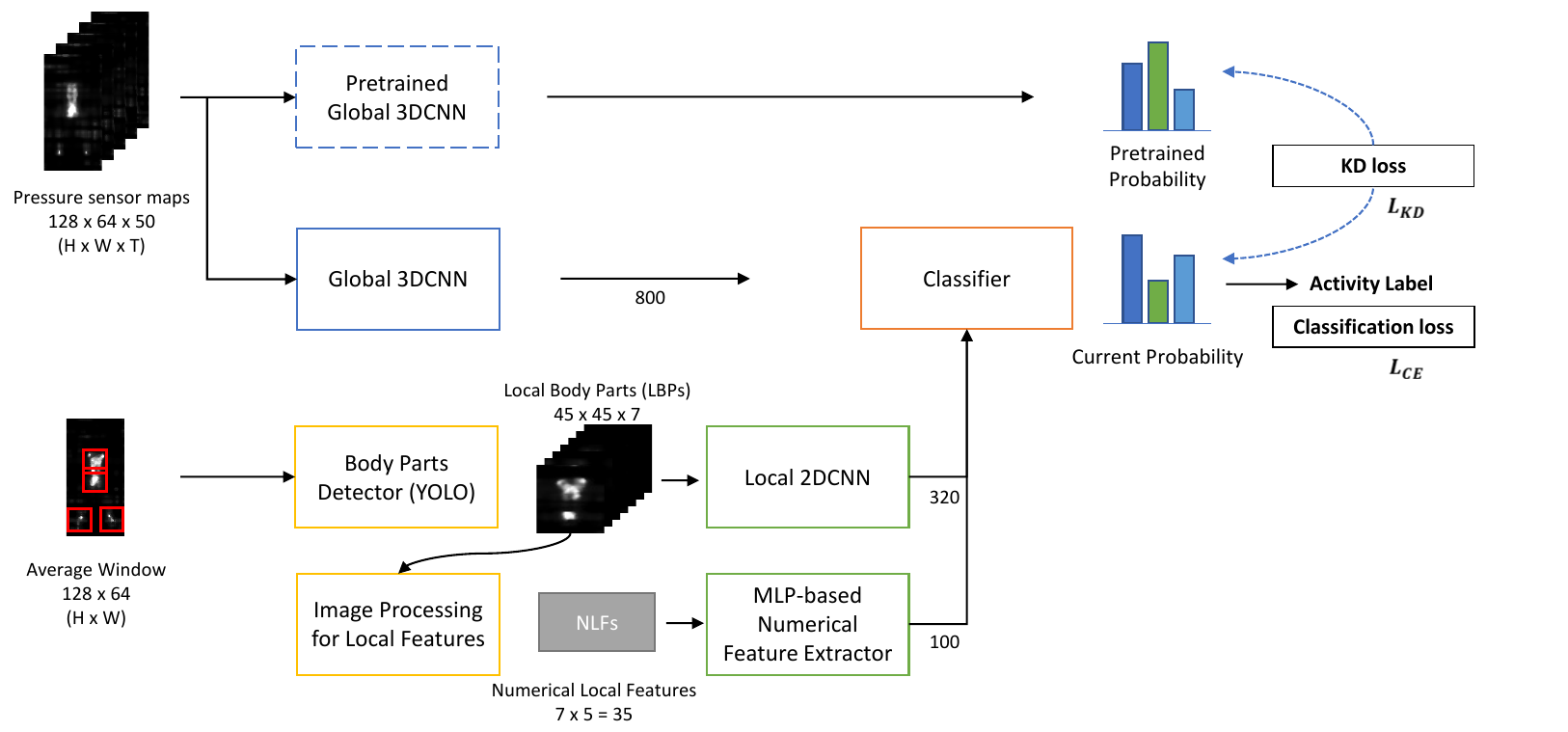}
        \vspace{-1cm}
        \end{center}
           \caption{The overview of the proposed architecture. The feature dimensions of the global 3DCNN, the local 2DCNN, and the MLP-based numerical feature extractor are 800, 320, and 100, respectively. $L_{KD}$ denotes the knowledge distillation loss between the predictions of student and teacher, and $L_{CE}$ the cross-entropy loss for ground truth and output of the student model.}
        \label{fig:short}
        \vspace{-2mm}
    \end{figure*}

    In this study, we propose a novel local-global feature fusion network architecture for recognizing body-weight exercises from pressure mapping sensors. While deep neural networks can distinguish human activities from pressure sensor maps, physical constraints on human activities can significantly improve the prediction of data-driven models. To address this, we use You Only Look Once (YOLO) to localize the human body part in the pressure sensor map and extract local features using a local CNN model. We also calculate numerical features such as orientations, areas, and angles of the body parts, using image processing techniques. These local and numerical features are embedded with the data-driven model to guide the training procedure as appropriate regularization. The global features are extracted from the global 3DCNN model, and we concatenate the global and local features into fully connected fusion layers to classify the body-weight exercises. To further improve the stability of the proposed network architecture training procedure and balance the optimization between the global data-driven model and local feature extraction models, we adopt the teacher-free knowledge distillation technique \cite{yuan2020revisiting}.
    
    
    The contributions of this paper can be summarized as follows:
    \begin{itemize}
        \vspace{-2mm}
        \item We propose a high-level local-global feature fusion framework for recognizing body-weight exercises with floor-based dynamic pressure maps.
        \item We detect and localize the different body parts using YOLO. 
        \item We adopt a knowledge distillation for regularization (KD-regularization) to keep the knowledge of the global feature extraction and improve the performance of body-weight exercise recognition.
        \item To demonstrate the effectiveness of the proposed architecture, we evaluate it on a dataset published in our previous work \cite{zhou2022quali}.
    \end{itemize}
    
    The rest of the paper is organized as follows. Section~\ref{sec:method} provides details of the proposed framework. Section~\ref{sec:experimentalresults} presents qualitative and quantitative experimental results. Finally, Section~\ref{sec:conclusion} concludes the paper and addresses future work.

\section{Proposed architecture}
\label{sec:method}

    We propose a local-global feature fusion network architecture for body-weight exercise recognition using pressure sensor maps. The goal of the proposed architecture is to extract high-level local features explicitly and leverage them along with global features for better activity recognition using knowledge distillation. The overall network architecture of the proposed method is illustrated in \cref{fig:short}. 
    Our study involves sequential steps, starting from the extraction of high-level local features to processing them using different neural network architectures and distilling the knowledge from global features to the local-global combined network architecture.
    In the first step, the proposed network architecture has three different networks: a local feature extractor, a global feature extractor, and a local numerical feature extractor. The local features, including the local numerical feature, are extracted from an average window of the corresponding activity. 
    In the second step, the local features are combined with the global feature, and the concatenated feature is fed into a multi-layer perceptron (MLP)-based classifier to classify body-weight exercises. To preserve the knowledge of global feature extraction and balance the optimization between the global and local feature extraction methods, we apply a knowledge distillation technique for regularization (KD-regularization).
    
    
    \subsection{Local Feature Extraction}

    We use YOLO to detect and localize body parts based on physical constraints. We classify the different body parts into three broad categories: terminal, torso, and limbs. We choose YOLO over a masked autoencoder model \cite{he2022masked} since the latter focuses more on image reconstruction, which is not aligned with our pressure map data in signal format. We finetune the YOLO detector to detect body parts on the pressure sensor maps and classify them. In any frame, the number of such pressure mappings differs since the maximum number of expected body pressure mapping is 7, where two limbs, one torso, and four terminals (2 hands and 2 feet). To make sure to get all the local information, we fixated the number of terminals to at most seven, and leveraged the pre-trained YOLO to extract these different pressure mappings.
    
    
    Two types of high-level local features are used in this study. The first type consists of pressure maps of different body parts stacked channels-wise in an image with each channel of dimension 45$\times$45. A maximum of seven total body parts are expected in any activity's average windows. If the number of detected body parts by YOLO is less than the reserved channels, it is replaced with an image of black pixels. This high-level local features dataset is named localized body parts (LBPs), where each sample results in (45$\times$45$\times$7) spatial dimensions. The second type is local numerical features (NLFs), extracted using image processing techniques on each channel of the previously created cropped image dataset. Five NLFs are extracted from each localized body part, including its location (x and y coordinates), area, and orientation (sine and cosine values). Hence, the LBPs with dimensions of 45$\times$45$\times$7 are obtained by YOLO for each average window, and the five NLFs are calculated for each body part.
    
    
    \subsection{Feature Fusion with Knowledge Distillation}
    
    Global features are extracted from the pressure-sensing map in the temporal domain and describe its overall outline, while local features focus on LBPs on the average window over the temporal domain. The 3DCNN processes the entire pressure sensing map to extract global features, the independent 2DCNN processes LBPs, and the MLP processes NLFs, as shown in \cref{fig:short}. To combine the global and local features, we use a stacked generalization ensemble method that learns how to combine predictions from multiple models \cite{brownlee2018better}.
    

    A 3x3 convolution filter processes 45x45 spatial images with seven channels.
    
    The training procedure for the proposed architecture consists of two steps. 
    First, we train the 3DCNN on the entire pressure sensor maps by using the cross-entropy for ground truth and model outputs as a preprocessing step.
    Then, we train the overall architecture, which includes the global 3DCNN, local 2DCNN, and MLP-based numerical feature extractor. 
    In the overall training procedure, we adopt the teach-free knowledge distillation technique \cite{yuan2020revisiting} to improve the stability of the training and control the importance of the feature combination.
    Generally, the knowledge distillation technique transfers knowledge from a teacher model to a smaller student model to improve the performance of the student model. In contrast, the teacher-free knowledge distillation aims to regularize the model to avoid overfitting. Likewise, in the proposed architecture, the pre-trained global 3DCNN serves as the teacher model, and the overall global-local feature combined network architecture is the student model. The output prediction of the pre-trained model is expressed as follows.
    \begin{equation}
        \label{eq:temperature}
        p^t_i(x;\tau) = softmax(z^t_i(x;\tau)) = \frac{\exp{(z^t_i(x)/\tau)}}{\sum_j \exp{(z^t_j(x)/\tau)}}
    \end{equation}
    where $z^t$ denotes the output logit vector of the pre-trained global CNN, and $\tau$ denotes a temperature parameter. 
    The idea of teacher-free knowledge distillation is to regularize the pre-trained and student models using the Kullback-Leibler (KL) divergence. We minimize the initial cross-entropy classification loss and the knowledge distillation regularization loss between the pre-trained and student model predictions.
    
    \begin{equation}
        \mathbb{L}_{KD} (X;\tau) = {KL} (p^t(x;\tau), p(x;\tau))
        \label{eq:kl}
    \end{equation}
    \begin{equation}
            L_{Final} = (1-\alpha) L_{CE}(q, p)+\alpha L_{KD}(X; \theta_{Pre}, \theta)
        \label{eq:selfkd}
    \end{equation}
    where $X$ is the input pressure sensor map data, $q$ is the ground truth, $p^t$ and $p$ are the output probability of the pretrained global 3DCNN and the overall student network, respectively, ${\tau}$ is the temperature, $\alpha$ is the weight, and $\theta$ and $\theta_{Pre}$ are the parameters of the overall global-local feature combined network and the pre-trained Global 3D CNN, respectively. $L_{CE}$ denotes the cross-entropy loss, $L_{KD}$ denotes the knowledge distillation loss, and $L_{Final}$ denotes the final objective loss to optimize the proposed network architecture.

\section{Experimental Results}
\label{sec:experimentalresults}
    In this section, we introduce the dataset and implementation details and provide the comparison results with various classification models, such as Temporal CNN \cite{azimi2020machine}, Transformer \cite{hou2022crack}, and TConv \cite{singh2017transforming}, introduced in \cite{zhou2022quali}, and conduct an ablation study. We also evaluate our work based on three recent instance segmentation models based on YOLO i.e.; YOLOv7 \cite{wang2023yolov7}, YOLOv8 \cite{dumitriu2023rip} and RT-DETR-I \cite{lv2023detrs}.   
    
    \textbf{Dataset:}
    We used the dataset published in the experiments in \cite{zhou2022quali}. It consists of spatial-temporal sequences of activities divided into nine categories. All the exercises chosen are body-weight exercises. Each category has variations of the exercise, and the dataset provides 47 variants of the exercises. Each category is well-distinguishable, but the variations are very similar. 
    For the collection of the dataset, twelve health participants performed these exercises on the pressure sensor mat. The participants are of different fitness levels, from expert to enthusiast. Out of 12 participants, four were female, and 8 were males. Their height varies from 162 cm to 189 cm. 

    
    \textbf{Implementation Details:}
    The proposed framework was implemented in PyTorch version 1.11.0 on Ubuntu 20.04.3 using NVIDIA RTX 3090 GPU. Adam optimizer was used to train the proposed architecture with an initial learning rate of 0.001 and a reduction of the learning rate by 0.1 on the plateau.

\begin{table}[!t]
\begin{center}
\caption{Comparison results with other methods and ablation study of the proposed method. The model marked * are trained on global features in study \cite{zhou2022quali}. The numbers are expressed in percent.}
\label{Tab:resulttab}
\small
\resizebox{\linewidth}{!}{
\begin{tabular}{|l|c|c|c|c|c|}
 \hline
 Model& KD regularization & Accuracy& Precision& Recall& F1 \\
 \hline
 TConv \cite{singh2017transforming} * &- & 48.7 & - & -  & -  \\
 Temporal CNN \cite{azimi2020machine} * &- & 51.2 & - & - & -   \\
 Transformer4D \cite{hou2022crack} * &-  & 52.9 & - & - & -   \\
 
 \hline
 
 Baseline   &- &63.3 & 63.1 & 62.7 & 62.9    \\
 OnlyLBP &- & 44.5 & 41.7 & 43.8 & 42.7    \\
 OnlyNLF &- & 20.5 & 12.1 & 19.7 & 14.9  \\
 NoLBP     &- &  64.3 & 59.9 & 63.7 & 61.7    \\
 NoLBP    &\checkmark &  65.2 & 64.6 & 64.8 & 64.7    \\
 NoNLF   &- & 66.0 & 65.8 & 65.2 & 65.4   \\
 NoNLF  &\checkmark & 66.3 & 65.9  & 65.3  & 65.6   \\
 Proposed (YOLOv7 \cite{wang2023yolov7}) & -  & 66.7 &67.1 & 66.2 & 66.7    \\
 Proposed (YOLOv7 \cite{wang2023yolov7})  &\checkmark  & 67.7 & 67.4 & 66.6&67.5\\
 \textbf{Proposed (YOLOv8 \cite{dumitriu2023rip})}  &\checkmark  & \textbf{69.2} & \textbf{70.5} & \textbf{78.3}& \textbf{73.9}\\
 Proposed (RT-DETR-l \cite{lv2023detrs})  &\checkmark  & 65.3 & 64.7 & 70.6 & 67.5 \\
\hline
\end{tabular}}
\end{center}

\vspace{-3mm}
\end{table} 

    
    \begin{figure}[!t]
        \centering
        \begin{subfigure}{\linewidth}
            \includegraphics[width=\linewidth]{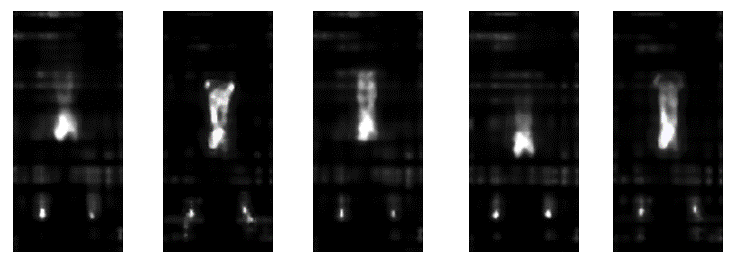}
            \vspace{-3mm}
            \caption{}
            \label{fig:catgory1}
        \end{subfigure}%
        \hfill
        \begin{subfigure}{\linewidth}
            \includegraphics[width=\linewidth]{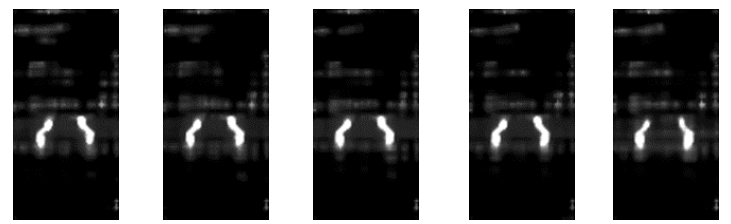}
            \caption{}
            \label{fig:catgory6}
            \vspace{-3mm}
        \end{subfigure}%
        \caption{Average window examples of pressure sensor maps of variants: (a) from Crunches exercise category, (b) from Standing Up category.}
        \label{fig:averageframe}
        \vspace{-5mm}
    \end{figure}
    
    
    \textbf{Results:}
    To evaluate our proposed architecture, we conducted comparison experiments with other methods and ablation study as shown in \cref{Tab:resulttab}. The proposed method outperformed other methods in terms of accuracy. To verify the effectiveness of the proposed local feature extraction methods and the KD-regularization, we derive five variants of the proposed architecture: ' Baseline', 'OnlyLBP', 'OnlyNLF', 'NoLBP', and 'NoNLF'. 'Baseline' is the global 3D CNN alone, 'OnlyLBP' is the 2DCNN alone, 'OnlyNLF' the MLP for the NLF alone, 'NoLBP' is the proposed architecture without LBPs, and 'NoNLF' is the proposed one without NLF, respectively. As shown in \cref{Tab:resulttab}, combining LBP and NLF with global features provided better results, and the proposed KD regularization improved the performance of the proposed architecture. 
    Also as a part of ablation studies, we experimented with three instance segmentation models derived from YOLO with YOLOv8 having the best performance.
    In \cref{fig:averageframe} (a), the intraclass differences are well observed in the average windows of different variants in the Crunches exercise category. The NLFs, such as the number and type of detections, size, and orientation, differ for each variant. Thus, the proposed framework can improve activity recognition for this category. 
    Without augmenting the number of training data, the proposed local-global feature framework could improve the performance. Thus, the effectiveness of the local-global feature framework is well-suited to tasks like activity recognition which is hard to get annotated data.
    
    \textbf{Limitations of proposed framework} As shown in \cref{fig:averageframe} (b), there is no significant difference in the average windows of variants in the Standing Up category. 
    Though we used the average window of the pressure sensor maps to detect enough LBPs for the input of the local feature extractions, the proposed framework cannot improve activity recognition performance compared to the baseline if there is no difference between the average windows.
    Some intra-categories lead to similar local features for different categories that do not significantly improve the proposed network over the baseline 3DCNN.
    
\section{Conclusion}
\label{sec:conclusion}
    This paper proposed a high-level local-global feature fusion network architecture for body-weight exercise recognition with a pressure mapping sensor mat. 
    The variants of exercises within the same category have very slight variations. To address those minor variations explicitly, we explore the local-global features combined architecture with the physical constraints for better activity recognition. This study explored that the local and global features combined with KD regularization can increase the quality of exercise recognition. Overall, all three proposed local-global feature combination models improved from the baseline. An improvement of 11.0 \% in terms of F1 score is observed between the baseline and proposed architecture. The proposed network architecture in this study can be applied to other studies where it is possible to extract the local features with physical constraints. In the future, we plan to utilize physics equations in neural networks to deeply consider the physical constraints and provide distinguishable local and global features.

\section{Acknowledgments}
\label{sec:ack}
The research reported in this paper was supported by the BMBF in the project VidGenSense (01IW21003) and Carl-Zeiss Stiftung under the Sustainable Embedded AI project (P2021-02-009). 

\newpage

\bibliographystyle{IEEEbib}
\bibliography{refs}

\end{document}